\documentclass[runningheads]{llncs}
\usepackage[T1]{fontenc}
\usepackage{amsfonts,amsmath}
\usepackage{graphicx}
\usepackage{array}
\usepackage{float}
\usepackage{tabularx,booktabs}
\usepackage{multirow,multicol}
\usepackage{subcaption}
\usepackage{booktabs}
\usepackage{longtable}
\usepackage{pdflscape}
\usepackage{array}

\newcommand{\mycomment}[1]{}

\usepackage{enumitem}
\setlist{nosep}                

\usepackage{etoolbox}
\apptocmd{\thebibliography}{\setlength{\itemsep}{0pt}\setlength{\parskip}{0pt}}{}{}

\begin{document}
\title{When to Adapt? \\ Adapting the Model or Data in Federated Medical Imaging}
%
%
\author{Chamani Shiranthika\orcidID{0000-0002-1608-8196} \and
Parvaneh Saeedi\orcidID{0000-0002-7507-9986} }
\authorrunning{C. Shiranthika et al.}
%
\institute{School of Engineering Science, Simon Fraser University, Burnaby, BC, V5A 1S6 
\email{{(csj5, psaeedi)}@sfu.ca}\\
}
\maketitle              
\begin{abstract}
Federated learning enables collaborative model training across medical institutions without sharing raw data, but its performance is often limited by domain heterogeneity across clients. Existing approaches to address this challenge fall into two main paradigms: model-side personalization, which adapts model parameters to each client, and data-side harmonization, which reduces inter-client variation at the input level. Despite their widespread use, these strategies have not been systematically compared. In this work, we conduct a comprehensive study across six medical imaging settings—colon polyp, skin lesion, and breast tumor segmentation, and tuberculosis CXR, brain tumor, and breast tumor classification—covering diverse types of domain shift. We evaluate a broad set of state-of-the-art harmonization and personalization methods under a unified framework. Our results reveal a conditional trade-off driven by the nature of heterogeneity: harmonization is more effective when variation is primarily appearance-based (e.g., CXR classification), while personalization performs better when differences are structural (e.g., colon polyp segmentation). When inter-client variation is limited, both strategies perform similarly. These findings demonstrate that the effectiveness of adaptation in federated medical imaging depends on the type and magnitude of domain shift rather than the strategy alone. We provide practical guidelines for selecting between harmonization and personalization, and highlight directions for future hybrid approaches that combine both paradigms. Code is available at: \url{https://github.com/ChamaniS/WhenToAdapt}.
\end{abstract}
\section{Introduction}
\label{sec:intro}
Federated learning (FL) \cite{McMahan_2017} has emerged as a key paradigm for privacy-preserving medical imaging collaborations, enabling multiple institutions to jointly train models without sharing patients' data. However, its effectiveness is often limited by inter-client heterogeneity, differences in data acquisition protocols, patient demographics, and annotation practices. Such non-independent and identically distributed (IID) data distributions cause severe model bias and poor cross-site generalization. Therefore, managing data heterogeneity has become one of the most critical open challenges in federated medical learning \cite{shiranthika2023decentralized}, \cite{shiranthika2024adaptive}.

Two broad research directions have evolved to address such heterogeneity challenge. \textit{(1): Model-side personalization} focuses on adapting the model to each client through modularized layers, local fine-tuning, or adaptive optimization. Early personalized federated learning methods rely on local fine-tuning or partial model sharing (FedPer \cite{arivazhagan2019federated}, FedRep \cite{collins2021exploiting}, FedBN \cite{li2021fedbn}), and on meta-learning or attention mechanisms (FedMeta \cite{chen2018federated}, pFedMe \cite{t2020personalized}, Per-FedAvg \cite{fallah2020personalized}, FedAMP \cite{huang2021personalized}). Some research treats clients as related tasks (MOCHA \cite{smith2017federated}, VIRTUAL \cite{corinzia2019variational}), clusters of similar clients (IFCA \cite{ghosh2020efficient}, FedCluster \cite{chen2020fedcluster}, FedGroup \cite{duan2021fedgroup}), or uses knowledge distillation (FedMD \cite{li2019fedmd}, FedDF \cite{zhang2025model}). Some other widely used methods include FedAdam \cite{reddi2020adaptive}, FedProx \cite{li2020federated}, SCAFFOLD \cite{karimireddy2020scaffold}, pFedMe \cite{t2020personalized}, FedMeta \cite{chen2018federated}, FedAMP \cite{huang2021personalized}, and Ditto \cite{li2021ditto}. \textit{(2): Data-side harmonization} reduces appearance discrepancies caused by scanner hardware, acquisition protocols, or imaging settings while preserving clinically relevant anatomy and is commonly studied as protocol-level preprocessing (resampling, intensity/colour normalization) or image-level harmonization using learned cross-domain mappings \cite{mali2021harmonization}. State-of-the-art (SoTA) data-side harmonization methods include GANs and related generative models (CycleGAN \cite{liu2023style}, CUT \cite{park2020contrastive}, StarGAN, DeepHarmony \cite{dewey2019deepharmony}, ImUnity \cite{cackowski2023imunity}, BlindHarmony \cite{jeong2023blindharmony}), attention/transformer-based approaches (AdaIN \cite{huang2017adain}, CoMoGAN \cite{pizzati2021comogan}, QS-Attn \cite{hu2022qsattn}, Harmonization Transformer \cite{guo2021image}), and newer diffusion-based models, which align input distributions across sites while maintaining diagnostic content.

Despite substantial progress, prior work has largely studied personalization and harmonization in isolation, without analyzing when each is preferable under different domain shifts. We address this gap with a systematic comparison across six medical imaging tasks: colon polyp, skin lesion, and breast tumor segmentation, and tuberculosis CXR, brain tumor, and breast tumor classification. These tasks span diverse heterogeneity types, from strong structural variation (e.g., colon polyp segmentation) to predominantly appearance-driven shifts (e.g., tuberculosis CXR classification), as well as moderate or mixed variations (e.g., skin, brain, and breast tumor tasks).

Using a unified experimental protocol, we evaluate 18 SoTA harmonization and personalization methods under identical federated settings. Our results reveal a conditional trade-off: harmonization is more effective for appearance-driven variation, while personalization performs better under strong structural heterogeneity; when variation is limited or mixed, both approaches perform similarly. We contribute: (1) a comprehensive comparison of harmonization and personalization in federated medical imaging, (2) an empirical analysis of how the type and magnitude of domain shift affect their effectiveness, and (3) a modular codebase supporting both paradigms. Section \ref{sec:methodology} describes the methodology, Section \ref{sec:experiments} presents the results, and Section \ref{sec:conclusions} concludes the paper.

\section{Methodology}
\label{sec:methodology}
\subsection{Problem Formulation}
We consider a cross-silo FL setting with $k$ medical institutions (clients), each holding a local dataset:
$D_k = \{(x_i^{(k)}, y_i^{(k)})\}_{i=1}^{n_k}$ drawn from an unknown distribution $\mathcal{P}_k$ over images $x$ and labels $y$.
A standard FL objective learns a single global model $f_\theta$ by minimizing the weighted empirical risk:
\begin{equation}
    \min_{\theta} \; \sum_{k=1}^K p_k \, 
    \mathbb{E}_{(x,y)\sim \mathcal{P}_k}\big[\ell(f_\theta(x), y)\big],
    \label{eq:global-objective}
\end{equation}
where $p_k = n_k / \sum_j n_j$ and $\ell$ is a task-appropriate loss. In practice, the client distributions $\{\mathcal{P}_k\}$ are heterogeneous. In medical imaging collaborations, this heterogeneity arises from two distinct, qualitatively different sources:

1. \textbf{Input-level/style-level shift} $\Delta_{\text{style}}$:  variations in scanner hardware, acquisition protocols, intensity scaling, contrast, noise level, padding, and background artifacts. These changes alter the global appearance statistics while largely preserving the anatomical layout and pathology of the data.

2. \textbf{Content-level/structural shift} $\Delta_{\text{content}}$:  variations in anatomy, pathology morphology and prevalence, viewpoint, endoscope/camera pose, and scene context. These differences alter the spatial structure and semantic content that the model must segment or classify.

Let $\Delta_k = (\Delta_{\text{style},k}, \Delta_{\text{content},k})$ denote the domain shift of client $k$. A single global model $f_\theta$ may underperform when $\Delta_k$ is large and heterogeneous across clients. We empirically compare these two families of  heterogeneity acting differently on:

\begin{itemize}
    \item \textbf{Data harmonization}: learn or apply client-specific transformations $T_k$ in image or feature space such that the transformed data $T_k(x)$ are statistically aligned to a reference domain, and then train a \emph{shared} model $f_\theta$ on harmonized data, and
    \item \textbf{Model personalization}: directly adapt the \emph{shared} model $f_\theta$ per client, yielding client-specific parameters $\theta_k$, while leaving the raw images unchanged.
\end{itemize}
Formally, a harmonization-based system learns
\begin{equation}
    \min_{\theta,\,\{T_k\}} \sum_{k=1}^K p_k \, 
    \mathbb{E}_{(x,y)\sim \mathcal{P}_k}\big[\ell(f_\theta(T_k(x)) y)\big]
    \label{eq:harmonization-objective}
\end{equation}
A personalization-based system instead learns client-specific models
\begin{equation}
    \min_{\{\theta_k\}} \sum_{k=1}^K p_k \, 
    \mathbb{E}_{(x,y)\sim \mathcal{P}_k}\big[\ell(f_{\theta_k}(x) y)\big]
    \label{eq:personalization-objective}
\end{equation}
Accordingly, our research question is:
\begin{quote}
    \emph{Under which types of domain shift does it pay to adapt the \textbf{data} (via harmonization), and under which types does it pay to adapt the \textbf{model} (via personalization)?}
\end{quote}

To answer this, we design two federated medical imaging benchmarks that are deliberately dominated by different heterogeneity types:  
(i) colon polyp segmentation with large structural/contextual variation across four endoscopy datasets and  
(ii) Tuberculosis CXR classification with largely preserved anatomy but strong scanner/intensity variation across four CXR datasets. 

\subsection{Model Personalization and data harmonization approaches employed}
We consider representative model personalization and data harmonization strategies commonly used in federated learning.

Model personalization methods adapt the global model to client-specific distributions. We group them into three categories: (1) layer-wise personalization, which shares a common backbone while keeping client-specific components local (e.g., FedPer \cite{arivazhagan2019federated}, FedRep \cite{collins2021exploiting}, FedBN \cite{li2021fedbn}); (2) optimization-based approaches, which modify the training objective or update rules to handle heterogeneity (e.g., FedProx \cite{li2020federated}, SCAFFOLD \cite{karimireddy2020scaffold}, FedAdam \cite{reddi2020adaptive}); and (3) bi-model and meta-learning approaches, which explicitly learn personalized client models alongside a global model (e.g., pFedMe \cite{dinh2020personalized}, Ditto \cite{li2021ditto}).

Data harmonization methods aim to reduce inter-client appearance variation while preserving underlying structure. We consider: (1) basic preprocessing and augmentation to standardize resolution and intensity; (2) histogram-based methods that align intensity distributions across clients; (3) frequency-domain adaptation (FDA) to transfer low-frequency style information; (4) MixStyle-based approaches that interpolate feature or input statistics to simulate domain shifts; and (5) generative translation methods (e.g., CycleGAN \cite{zhu2017unpaired} , CUT \cite{park2020contrastive}, AdaIN \cite{huang2017adain}, CoMoGAN \cite{pizzati2021comogan} ) that learn mappings between domains.

\begin{figure}
\centerline{\includegraphics[scale = 0.5]{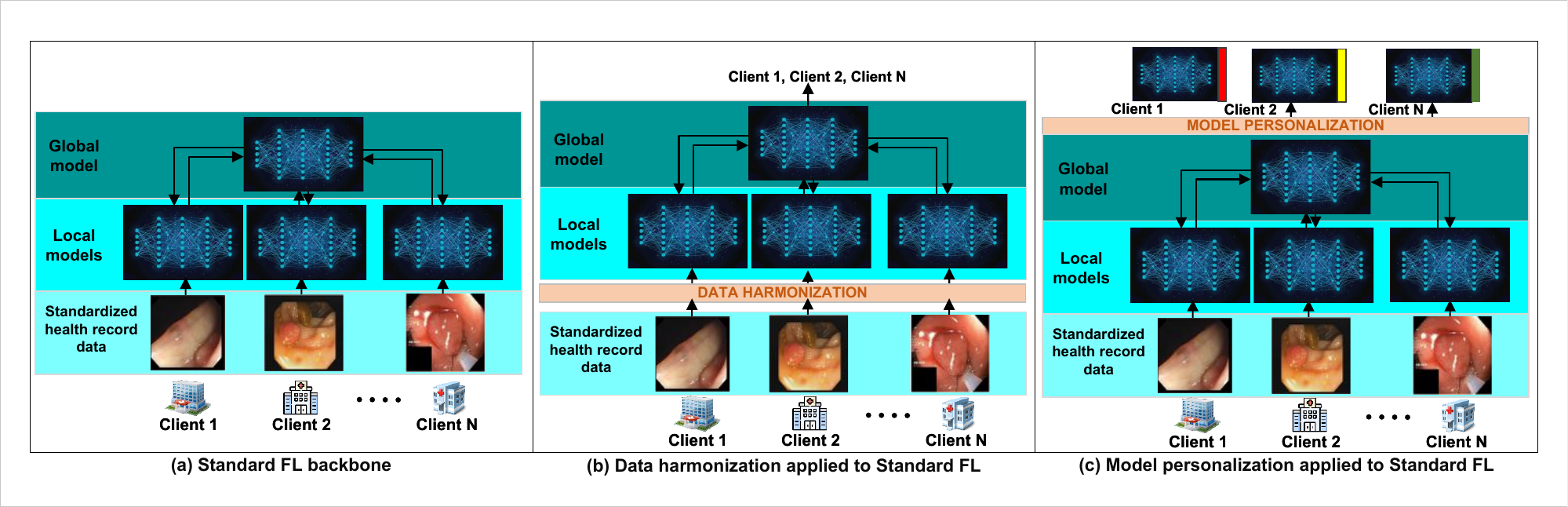}}
\caption{\scriptsize Experimental methodology of the study.}
\label{fig:whentoadaptmeth}
\end{figure}
\subsection{Comparative Framework}
As shown in Fig. \ref{fig:whentoadaptmeth}, we instantiate model personalization and data  harmonization strategies on top of a standard FL backbone (Flower \cite{beutel2020flower}-based cross-silo FL with four clients per task). Each method uses the same base architecture for each task: a U-Net for colon polyp, skin lesion, and breast tumor segmentation; a DenseNet-based classifier for tuberculosis CXR; and an EfficientNet-based classifier for brain and breast tumor classification. All methods are trained with the same schedule and communication budget, ensuring that performance differences arise solely from the applied personalization or harmonization strategy.

\emph{Tasks. }We design six FL settings using medical imaging tasks shown in Table \ref{tab:dataset_summary}:

\emph{Baselines.}
We include five non-personalized and non-harmonized baselines:
\begin{itemize}
    \item \textbf{Locally-centralized}: each client trains its model solely on local data.
    \item \textbf{Centralized-globally tested}: a central model is trained on all clients' data and tested using all clients' pooled data.
    \item \textbf{Centralized-locally tested}: a central model is trained on all clients' data and tested using clients' local data separately in their local spaces.    
    \item \textbf{Vanilla FedAvg-globally tested}: a single global model trained by standard FedAvg and tested using all clients' pooled data.
    \item \textbf{Vanilla FedAvg-locally tested}: a single global model trained by standard FedAvg and tested using clients' local data separately in their local spaces. 
\end{itemize}

\begin{table*}[t]
\centering
\tiny
\caption{Dataset composition across clients for the 6 tasks. Each client corresponds to a dataset with its sample size and fixed test split. Remaining data is split into 85\% training and 15\% validation.}
\label{tab:dataset_summary}
\setlength{\tabcolsep}{0.2pt}
\renewcommand{\arraystretch}{1.1}
\begin{subtable}[t]{0.3\textwidth}
\centering
\caption{Segmentation tasks}
\begin{tabular}{@{} l l l c c @{}}
\toprule
\textbf{C\#} & \textbf{Dataset} & \textbf{Modality} & \textbf{Total \#} & \textbf{Test \#} \\
\midrule
\multicolumn{5}{c}{\textit{Task 1: Colon polyp segmentation}} \\ \midrule
C1 & Kvasir-SEG \cite{jha2019kvasirseg} & Endoscopy & 1000 & 100 \\
C2 & ETIS-Larib \cite{silva2014toward} & Endoscopy & 196 & 19 \\
C3 & CVC-ColonDB \cite{bernal2012towards} & Endoscopy & 380 & 38 \\
C4 & CVC-ClinicDB \cite{bernal2015wmdova} & Endoscopy & 612 & 60 \\
\midrule
\multicolumn{5}{c}{\textit{Task 2: Skin lesion segmentation}} \\ \midrule
C1 & HAM10K \cite{tschandl2018ham10000} & Dermoscopy & 10015 & 1001 \\
C2 & PH2 \cite{mendona2013ph2} & Dermoscopy & 200 & 20 \\
C3 & ISIC2017 \cite{codella2017skin} & Dermoscopy & 2000 & 200 \\
C4 & ISIC2018 \cite{codella2018isic} & Dermoscopy & 2594 & 259 \\
\midrule
\multicolumn{5}{c}{\textit{Task 3: Breast tumor segmentation}} \\ \midrule
C1 & BUSBRA \cite{gomez2024bus} & Ultrasound & 1875 & 81 \\
C2 & BUS\_UC \cite{busuc_dataset} & Ultrasound & 811 & 187 \\
C3 & BUSI \cite{al2019dataset} & Ultrasound & 437 & 43 \\
C4 & UDIAT \cite{yap2018udiat} & Ultrasound & 163 & 16 \\
\bottomrule
\end{tabular}
\end{subtable}
\hfill
\begin{subtable}[t]{0.55\textwidth}
\centering
\caption{Classification tasks}
\begin{tabular}{@{} l l l c c @{}}
\toprule
\textbf{C\#} & \textbf{Dataset} & \textbf{Modality} & \textbf{Total \#} & \textbf{Test \#} \\
\midrule
\multicolumn{5}{c}{\textit{Task 4: Tuberculosis CXR classification}} \\ \midrule
C1 & Shenzhen Hospital CXR \cite{jaeger2014two} & X-ray & 662 & 65 \\
C2 & Montgomery County CXR \cite{jaeger2014two} & X-ray & 138 & 13 \\
C3 & TBX11K \cite{Liu_2020_CVPR} & X-ray & 1600 & 160 \\
C4 & Pakistan (local) & X-ray & 3008 & 300 \\
\midrule
\multicolumn{5}{c}{\textit{Task 5: Brain tumor classification}} \\ \midrule
C1 & Sartajbhuvaji (Kaggle) \cite{bhuvaji2020brain} & MRI & 3160 & 314 \\
C2 & RM1000 (Kaggle) \cite{rm1000_brain_mri} & MRI & 7023 & 701 \\
C3 & thomasdubail (Kaggle) \cite{dubail2020brain256} & MRI & 3096 & 308 \\
C4 & Figshare \cite{cheng2017brain} & MRI & 7200 & 720 \\
\midrule
\multicolumn{5}{c}{\textit{Task 6: Breast tumor classification}} \\ \midrule
C1 & BUSBRA \cite{gomez2024bus} & Ultrasound & 1875 & 186 \\
C2 & BUS\_UC \cite{busuc_dataset} & Ultrasound & 811 & 80 \\
C3 & BUSI \cite{al2019dataset} & Ultrasound & 647 & 64 \\
C4 & UDIAT \cite{yap2018udiat} & Ultrasound & 163 & 15 \\
\bottomrule
\end{tabular}
\end{subtable}
\end{table*}
\section{Experimental Results}
\label{sec:experiments}
Table \ref{tab:combined_harmonization} shows one representative colon polyp sample per dataset after harmonization. Harmonized samples of other datasets are included in our code repository and are omitted here for space. We include the amplified difference, defined as the absolute per-pixel RGB difference between the original and harmonized images, scaled to highlight subtle color and contrast changes. Segmentation tasks use Dice as the primary metric (with IoU, pixel accuracy, precision, recall, and specificity), while classification tasks use Cohen’s $\kappa$ (with accuracy, precision, recall, F1, and specificity). We compare harmonization and personalization methods and analyze trends by heterogeneity type (style vs.\ content). Results are reported in Tables \ref{tab:compound_subtables} and \ref{tab:compound_subtables_classification}.
\begin{table}[H]
\centering
\caption{ Examples of harmonized samples and amplified differences for Colon polyp (top), and skin lesion (bottom) datasets used for segmentation. H. = harmonized image; D. = amplified difference; Aug. = augmentation; Hist.mat. = histogram matching; ARI = average representative image; SRI = single representative image; FDA = Fourier domain adaptation.}
\label{tab:combined_harmonization}
\scriptsize
\renewcommand{\arraystretch}{1.5}
\setlength{\tabcolsep}{2pt}
\begin{tabular}{|c|l|c|c|c|c|c|c|c|c|c|c|}
\hline
\textbf{Image} & & \textbf{Aug.} & \textbf{ \shortstack{Hist.\\mat.\\(SRI)}}
& \textbf{\shortstack{Hist.\\mat.\\(ARI)}} & \textbf{FDA}
& \textbf{\shortstack{Input\\level\\mixstyle}}
& \textbf{\shortstack{Feature\\level\\mixstyle}}
& \textbf{\shortstack{Cycle-\\GAN}}
& \textbf{CUT} & \textbf{AdaIN} & \textbf{\shortstack{CoMo-\\GAN}} \\
\hline
\multicolumn{12}{|c|}{\textbf{Colon polyp datasets}} \\ \hline
\multirow{2}{*}{\begin{tabular}{c} \textbf{\small{C1}}\\
\includegraphics[width=0.30in,height=0.30in]{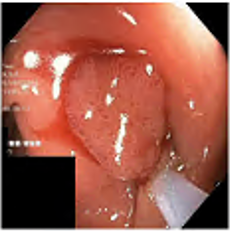}\end{tabular}}
& \textbf{H.}
& \includegraphics[width=0.30in,height=0.30in]{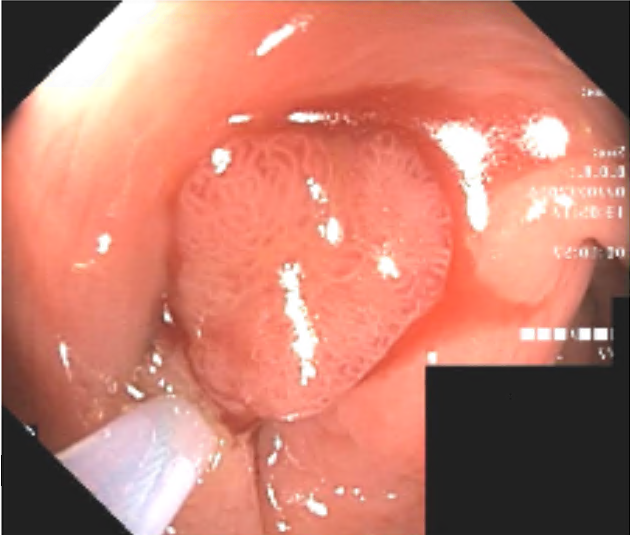}
& \includegraphics[width=0.30in,height=0.30in]{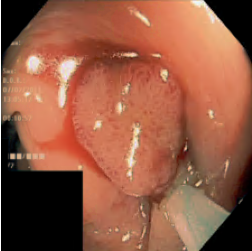}
& \includegraphics[width=0.30in,height=0.30in]{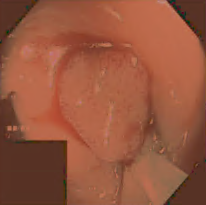}
& \includegraphics[width=0.30in,height=0.30in]{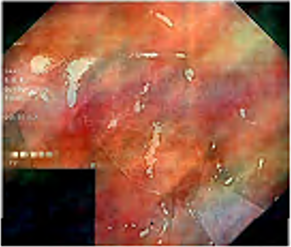}
& \includegraphics[width=0.30in,height=0.30in]{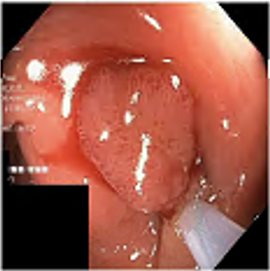}
& \includegraphics[width=0.30in,height=0.30in]{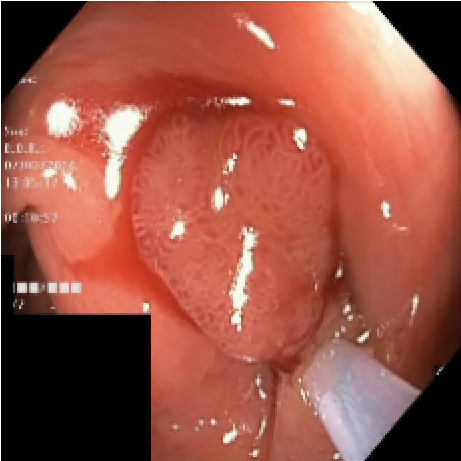}
& \includegraphics[width=0.30in,height=0.30in]{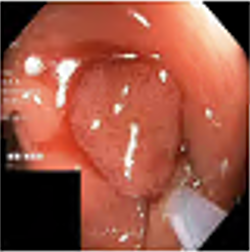}
& \includegraphics[width=0.30in,height=0.30in]{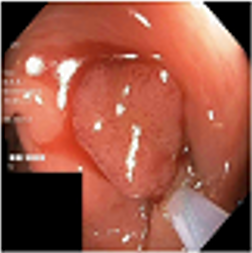}
& \includegraphics[width=0.30in,height=0.30in]{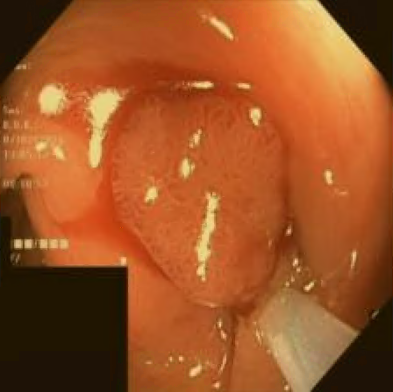}
& \includegraphics[width=0.30in,height=0.30in]{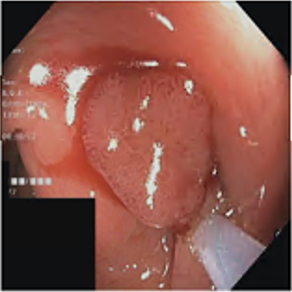} \\
\cline{2-12}
& \textbf{D.}
& \includegraphics[width=0.30in,height=0.30in]{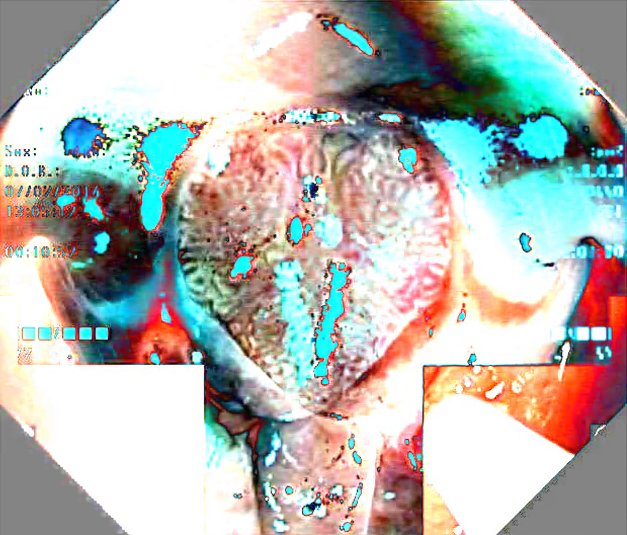}
& \includegraphics[width=0.30in,height=0.30in]{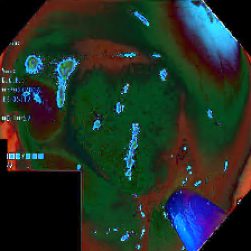}
& \includegraphics[width=0.30in,height=0.30in]{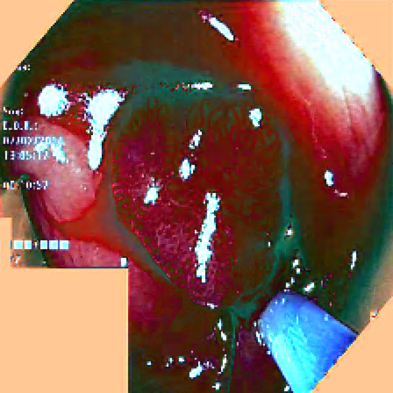}
& \includegraphics[width=0.30in,height=0.30in]{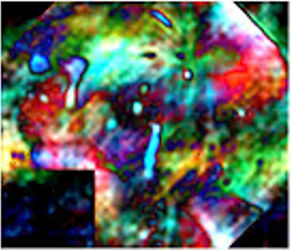}
& \includegraphics[width=0.30in,height=0.30in]{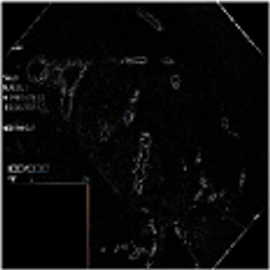}
& \includegraphics[width=0.30in,height=0.30in]{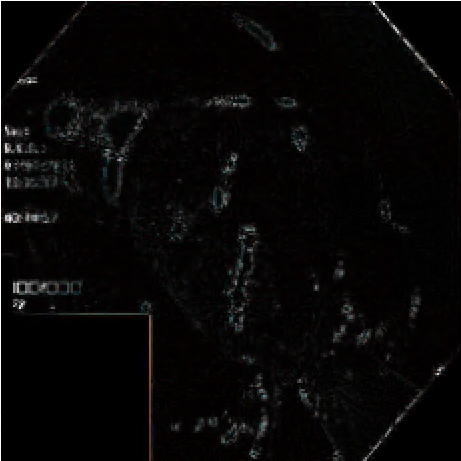}
& \includegraphics[width=0.30in,height=0.30in]{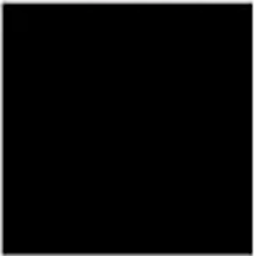}
& \includegraphics[width=0.30in,height=0.30in]{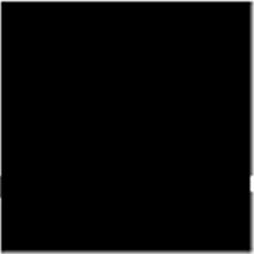}
& \includegraphics[width=0.30in,height=0.30in]{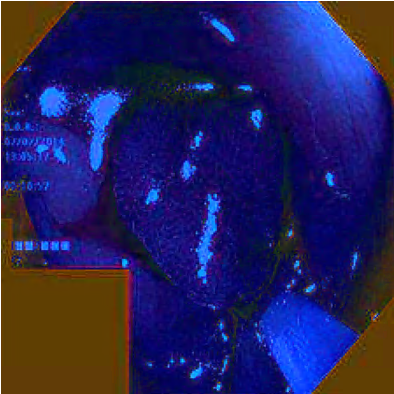}
& \includegraphics[width=0.30in,height=0.30in]{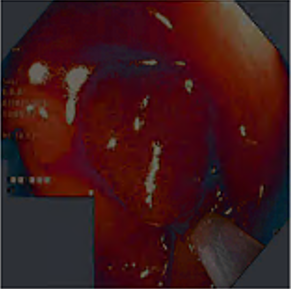} \\
\hline

\multirow{2}{*}{\begin{tabular}{c}  \textbf{\small{C2}}\\ \includegraphics[width=0.30in,height=0.30in]{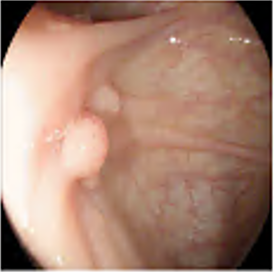}\end{tabular}}
& \textbf{H.}
& \includegraphics[width=0.30in,height=0.30in]{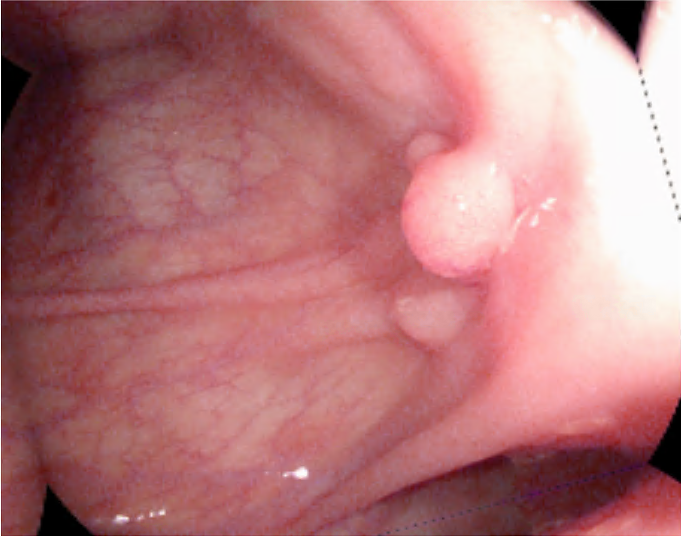}
& \includegraphics[width=0.30in,height=0.30in]{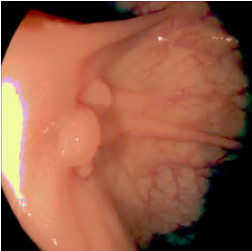}
& \includegraphics[width=0.30in,height=0.30in]{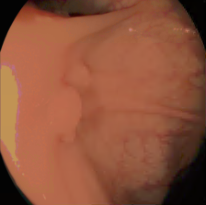}
& \includegraphics[width=0.30in,height=0.30in]{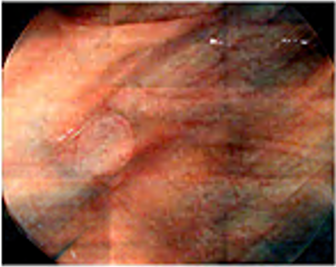}
& \includegraphics[width=0.30in,height=0.30in]{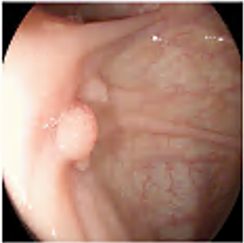}
& \includegraphics[width=0.30in,height=0.30in]{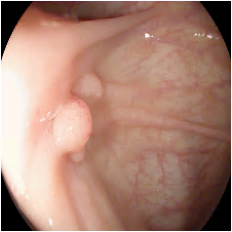}
& \includegraphics[width=0.30in,height=0.30in]{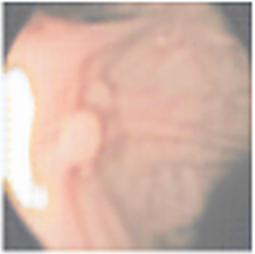}
& \includegraphics[width=0.30in,height=0.30in]{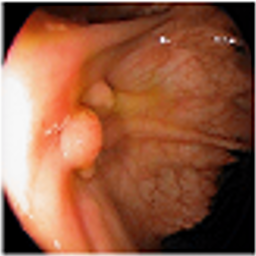}
& \includegraphics[width=0.30in,height=0.30in]{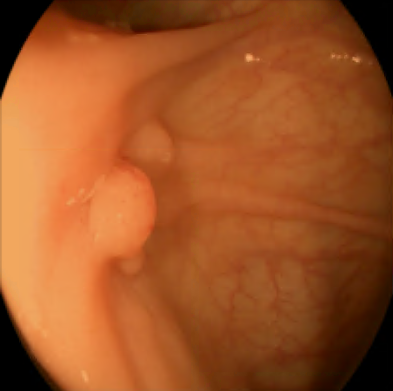}
& \includegraphics[width=0.30in,height=0.30in]{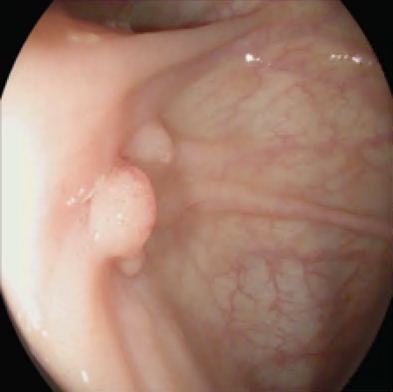} \\
\cline{2-12}
& \textbf{D.}
& \includegraphics[width=0.30in,height=0.30in]{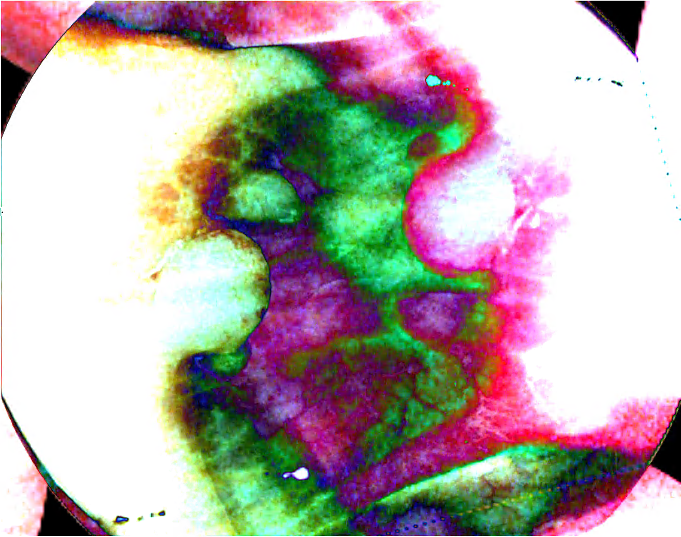}
& \includegraphics[width=0.30in,height=0.30in]{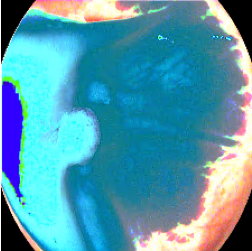}
& \includegraphics[width=0.30in,height=0.30in]{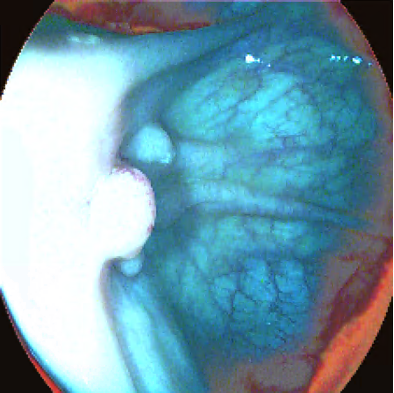}
& \includegraphics[width=0.30in,height=0.30in]{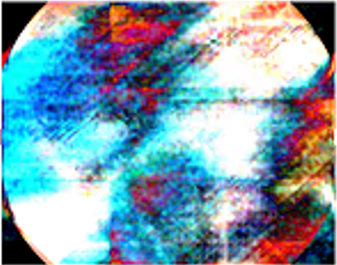}
& \includegraphics[width=0.30in,height=0.30in]{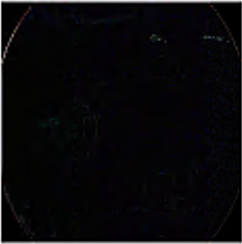}
& \includegraphics[width=0.30in,height=0.30in]{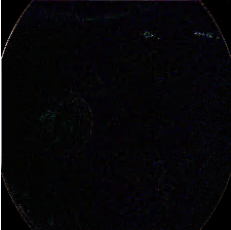}
& \includegraphics[width=0.30in,height=0.30in]{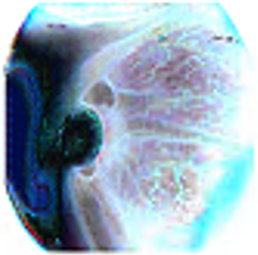}
& \includegraphics[width=0.30in,height=0.30in]{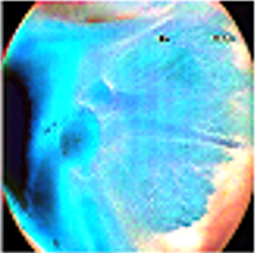}
& \includegraphics[width=0.30in,height=0.30in]{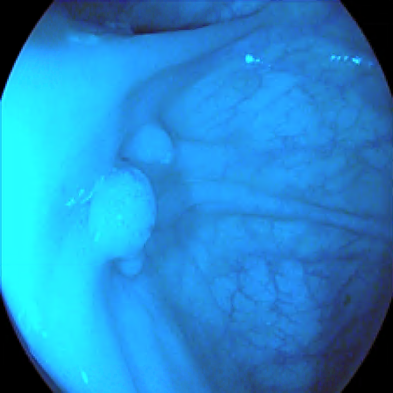}
& \includegraphics[width=0.30in,height=0.30in]{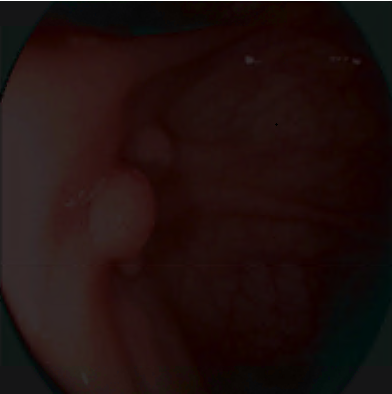} \\
\hline

\multirow{2}{*}{\begin{tabular}{c} \textbf{\small{C3}}\\  \includegraphics[width=0.30in,height=0.30in]{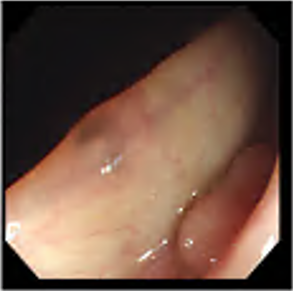}\end{tabular}}
& \textbf{H.}
& \includegraphics[width=0.30in,height=0.30in]{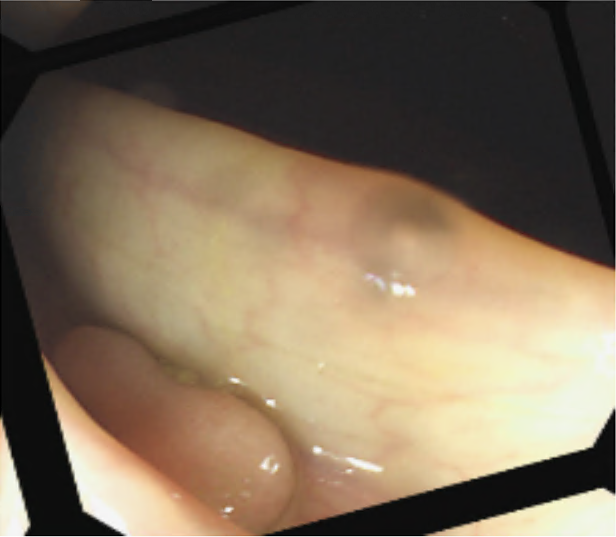}
& \includegraphics[width=0.30in,height=0.30in]{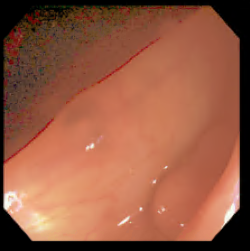}
& \includegraphics[width=0.30in,height=0.30in]{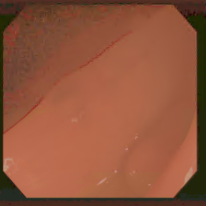}
& \includegraphics[width=0.30in,height=0.30in]{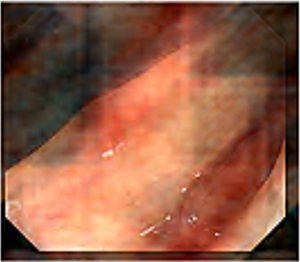}
& \includegraphics[width=0.30in,height=0.30in]{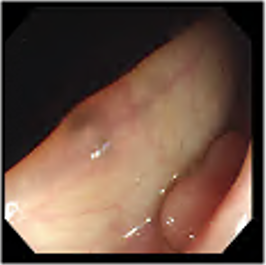}
& \includegraphics[width=0.30in,height=0.30in]{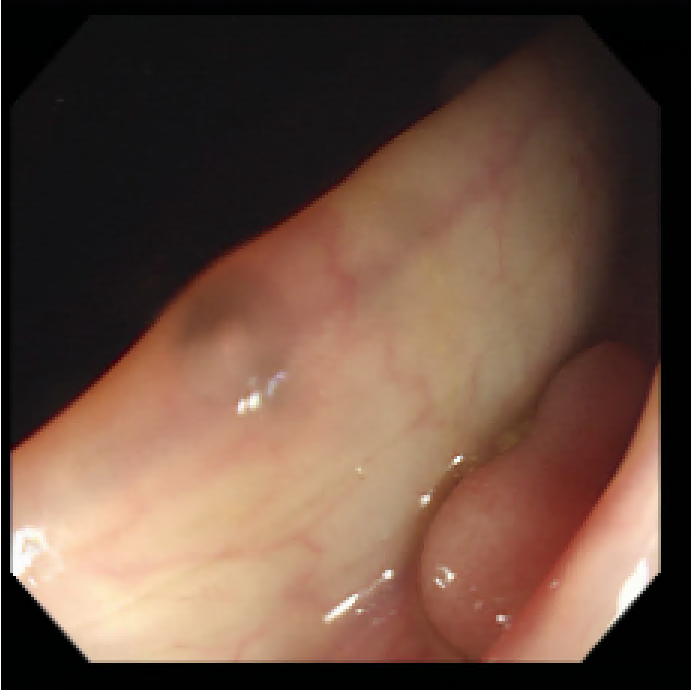}
& \includegraphics[width=0.30in,height=0.30in]{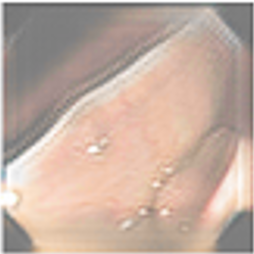}
& \includegraphics[width=0.30in,height=0.30in]{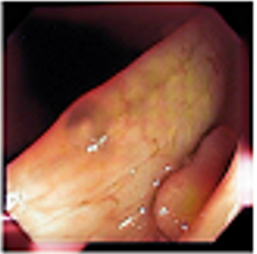}
& \includegraphics[width=0.30in,height=0.30in]{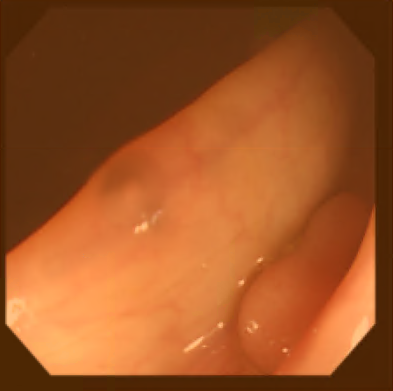}
& \includegraphics[width=0.30in,height=0.30in]{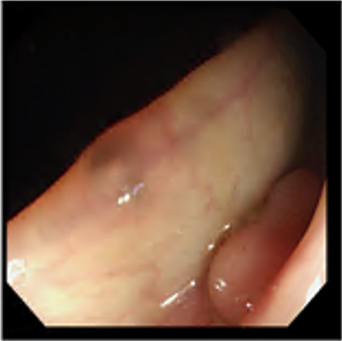} \\
\cline{2-12}
& \textbf{D.}
& \includegraphics[width=0.30in,height=0.30in]{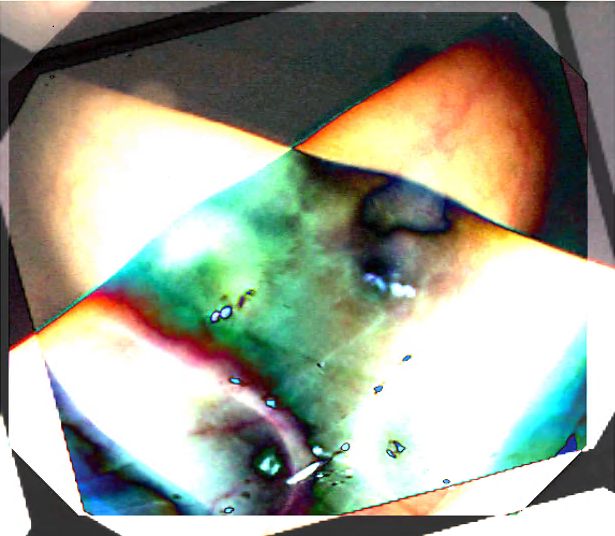}
& \includegraphics[width=0.30in,height=0.30in]{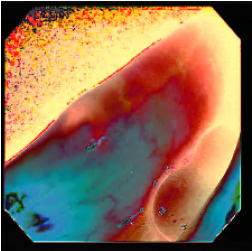}
& \includegraphics[width=0.30in,height=0.30in]{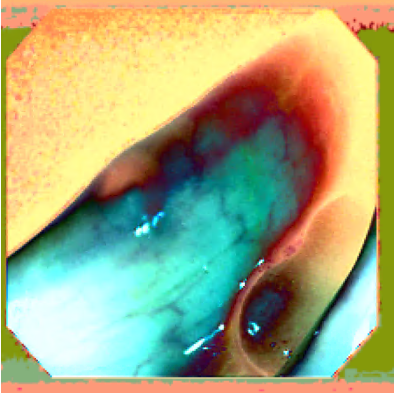}
& \includegraphics[width=0.30in,height=0.30in]{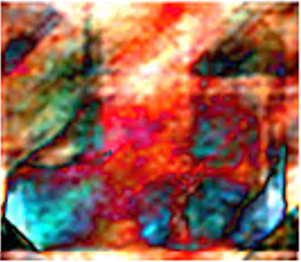}
& \includegraphics[width=0.30in,height=0.30in]{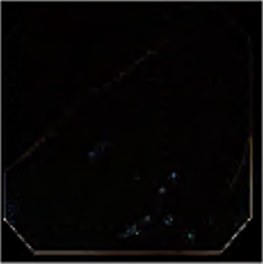}
& \includegraphics[width=0.30in,height=0.30in]{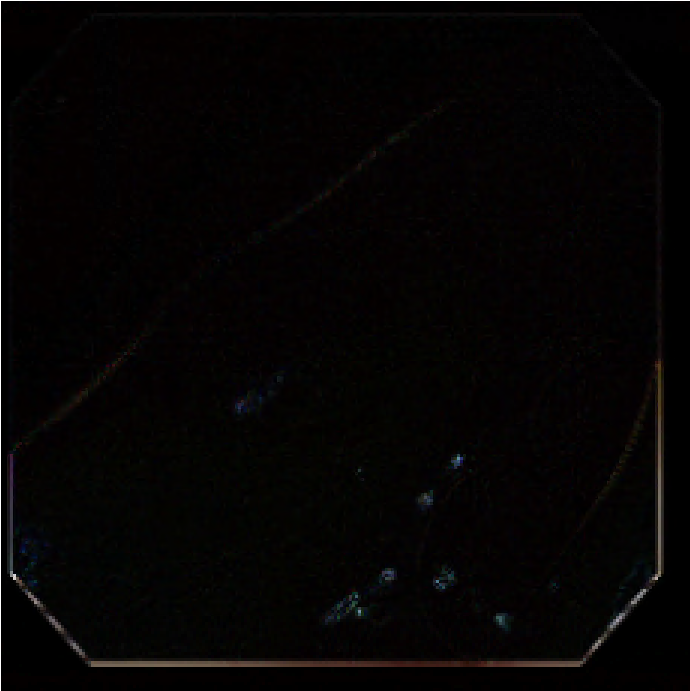}
& \includegraphics[width=0.30in,height=0.30in]{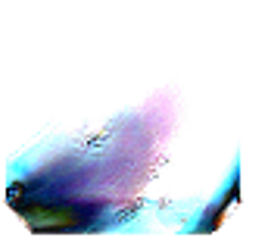}
& \includegraphics[width=0.30in,height=0.30in]{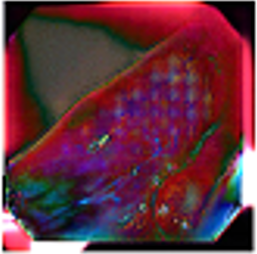}
& \includegraphics[width=0.30in,height=0.30in]{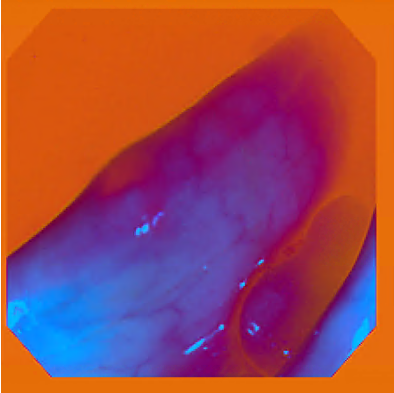}
& \includegraphics[width=0.30in,height=0.30in]{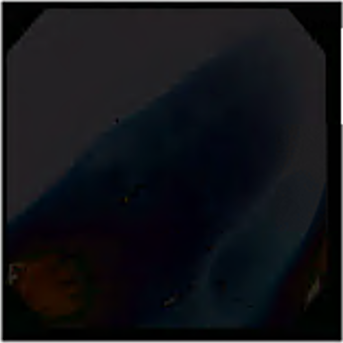} \\
\hline

\multirow{2}{*}{\begin{tabular}{c} \textbf{\small{C4}}\\  \includegraphics[width=0.30in,height=0.30in]{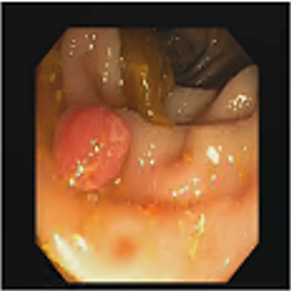}\end{tabular}}
& \textbf{H.}
& \includegraphics[width=0.30in,height=0.30in]{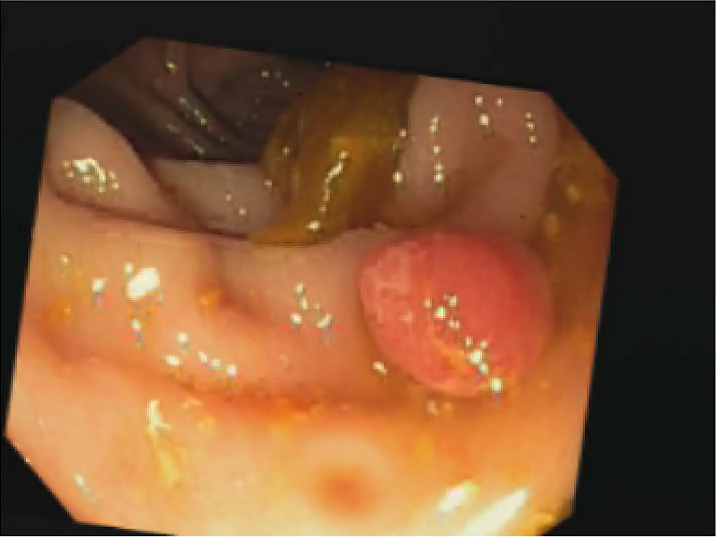}
& \includegraphics[width=0.30in,height=0.30in]{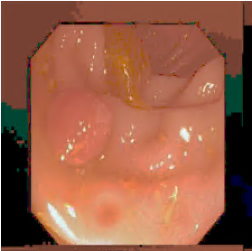}
& \includegraphics[width=0.30in,height=0.30in]{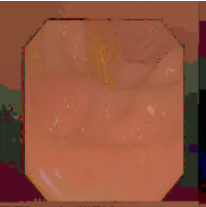}
& \includegraphics[width=0.30in,height=0.30in]{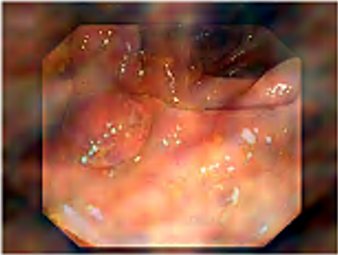}
& \includegraphics[width=0.30in,height=0.30in]{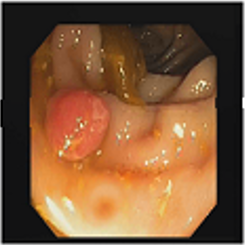}
& \includegraphics[width=0.30in,height=0.30in]{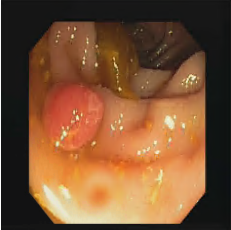}
& \includegraphics[width=0.30in,height=0.30in]{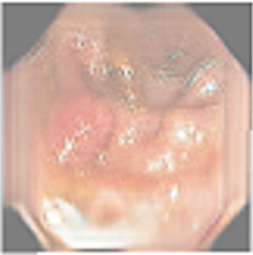}
& \includegraphics[width=0.30in,height=0.30in]{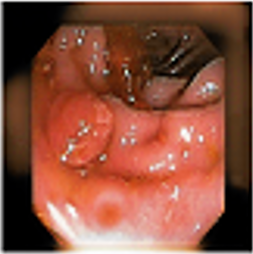}
& \includegraphics[width=0.30in,height=0.30in]{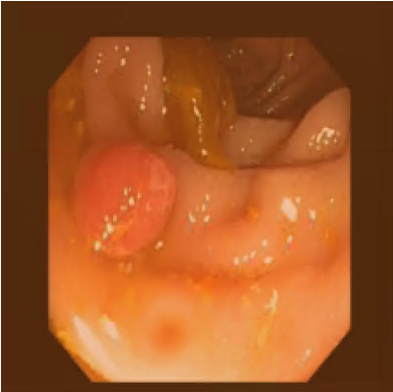}
& \includegraphics[width=0.30in,height=0.30in]{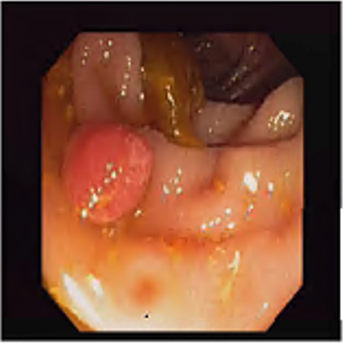} \\
\cline{2-12}
& \textbf{D.}
& \includegraphics[width=0.30in,height=0.30in]{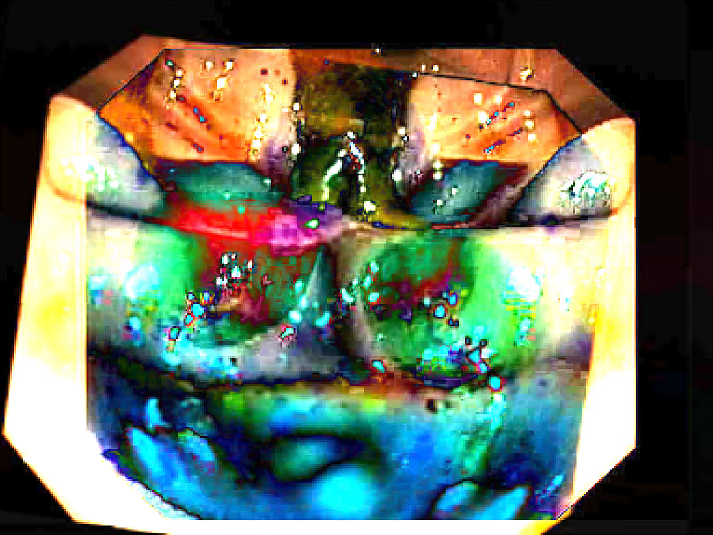}
& \includegraphics[width=0.30in,height=0.30in]{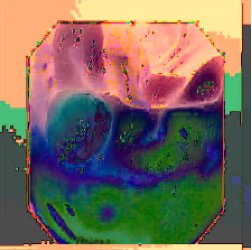}
& \includegraphics[width=0.30in,height=0.30in]{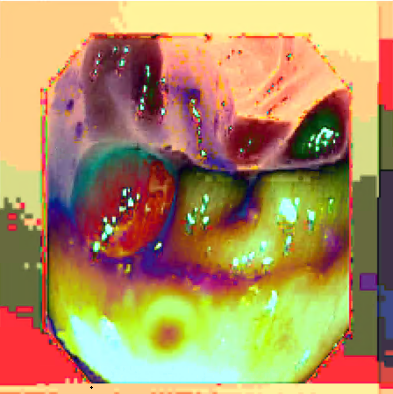}
& \includegraphics[width=0.30in,height=0.30in]{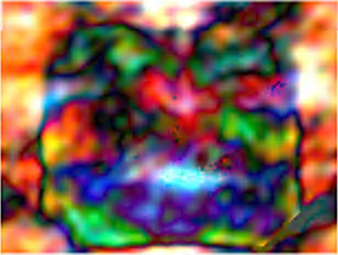}
& \includegraphics[width=0.30in,height=0.30in]{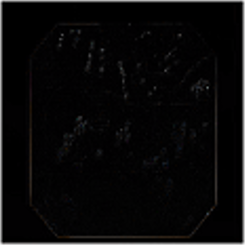}
& \includegraphics[width=0.30in,height=0.30in]{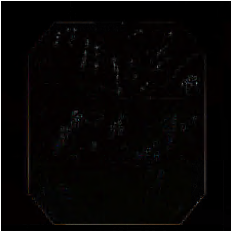}
& \includegraphics[width=0.30in,height=0.30in]{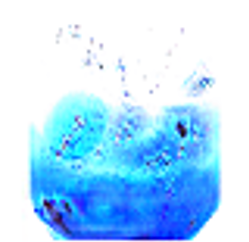}
& \includegraphics[width=0.30in,height=0.30in]{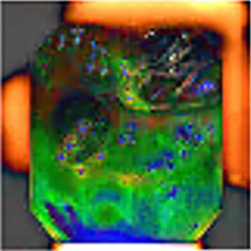}
& \includegraphics[width=0.30in,height=0.30in]{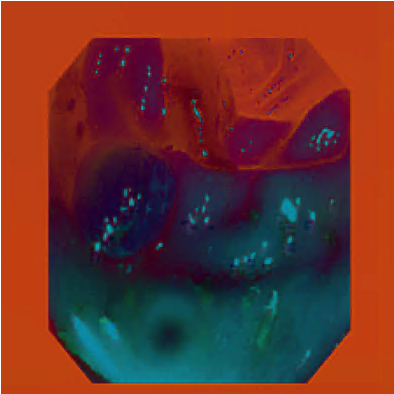}
& \includegraphics[width=0.30in,height=0.30in]{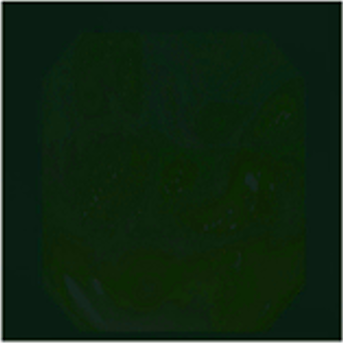} \\
\hline
\end{tabular}
\end{table}

\begin{table*}[t]
\tiny
\caption{Data harmonization vs. model personalization across segmentation tasks. ARI = average representative image, SRI = single representative image. Best results are in \textbf{bold}.}
\centering
\setlength{\tabcolsep}{2pt}
\renewcommand{\arraystretch}{1.2}

\begin{tabular}{@{} l | cccc | cccc | cccc @{}}
\toprule
& \multicolumn{4}{c|}{\textbf{Colon Polyp}} 
& \multicolumn{4}{c|}{\textbf{Skin Lesion}} 
& \multicolumn{4}{c}{\textbf{Breast Tumor}} \\

\textbf{Method} 
& C1 & C2 & C3 & C4 
& C1 & C2 & C3 & C4 
& C1 & C2 & C3 & C4 \\
\midrule

\multicolumn{13}{c}{\textit{Baseline methods}} \\ \midrule

Locally centralized
& 0.8297 & 0.7804 & 0.8857 & 0.8500
& 0.8920 & 0.8073 & 0.7692 & 0.8071
& 0.8367 & 0.9135 & 0.8049 & 0.7325 \\

Centralized -- globally tested
& \multicolumn{4}{c}{0.8499}
& \multicolumn{4}{c}{0.9250}
& \multicolumn{4}{c}{0.9008} \\

Centralized -- locally tested
& 0.8024 & 0.7892 & 0.9121 & 0.8428
& 0.9178 & 0.8928 & 0.9374 & 0.9434
& 0.8998 & 0.9307 & 0.8419 & 0.9153 \\

Vanilla FedAvg -- globally tested
& \multicolumn{4}{c}{0.7286}
& \multicolumn{4}{c}{0.9355}
& \multicolumn{4}{c}{0.8933} \\

Vanilla FedAvg -- locally tested
& 0.7559 & 0.4430 & 0.7508 & 0.8258
& 0.9283 & 0.9009 & 0.9448 & 0.9476
& 0.8964 & 0.9316 & 0.8027 & 0.9129 \\

\midrule
\multicolumn{13}{c}{\textit{Harmonization methods}} \\ \midrule

Augmentations
& 0.6698 & 0.3014 & 0.4714 & 0.6804
& 0.9217 & 0.8800 & 0.9322 & 0.9394
& 0.9351 & 0.9178 & 0.7424 & 0.9220 \\

Histogram matching (ARI)
& 0.7149 & 0.3901 & 0.5892 & 0.7144
& 0.9153 & 0.8909 & 0.9322 & 0.9379
& 0.9145 & 0.8936 & 0.7997 & 0.8875 \\

Histogram matching (SRI)
& 0.7749 & 0.6626 & 0.7336 & 0.7733
& 0.9141 & 0.9162 & 0.9307 & 0.9374
& 0.9193 & 0.8989 & 0.7747 & 0.8937 \\

FDA (SRI)
& 0.7498 & 0.3888 & 0.6928 & 0.7233
& 0.8940 & 0.8465 & 0.9176 & 0.9216
& 0.8592 & 0.8807 & 0.7842 & 0.8477 \\

Input level MixStyle
& 0.8296 & 0.6821 & 0.8270 & 0.8428
& 0.9114 & 0.8672 & 0.9324 & 0.9374
& 0.9173 & 0.8990 & 0.8073 & 0.9126 \\

Feature level MixStyle
& 0.8336 & 0.6418 & 0.8273 & 0.8324
& 0.9237 & 0.8894 & 0.9375 & 0.9428
& 0.9237 & 0.8992 & 0.8022 & 0.9026 \\

Cycle-GAN
& 0.8129 & 0.5506 & 0.6435 & 0.6685
& 0.9310 & 0.8926 & 0.9176 & 0.9355
& 0.8656 & 0.9174 & 0.7836 & 0.8390 \\

CUT
& 0.6918 & 0.6069 & 0.8222 & 0.7056
& 0.9078 & 0.5536 & 0.4683 & 0.7853
& 0.5758 & 0.8600 & 0.3547 & 0.0421 \\

AdaIN
& 0.7062 & 0.6282 & 0.8676 & 0.7679
& 0.7666 & 0.6694 & 0.7564 & 0.7484
& 0.9163 & 0.9025 & 0.8156 & 0.9062 \\

ComoGAN
& 0.7774 & 0.5653 & 0.8498 & 0.8418
& 0.9204 & 0.9086 & 0.9362 & 0.9420
& 0.9185 & 0.9033 & 0.8014 & 0.9273 \\

\midrule
\multicolumn{13}{c}{\textit{Personalization methods}} \\ \midrule

Local Finetuning
& 0.8329 & 0.6914 & 0.8678 & 0.8311
& 0.9290 & 0.9100 & 0.9513 & 0.9501
& 0.9195 & 0.9087 & 0.8014 & 0.8875 \\

FedPer
& 0.8085 & 0.4808 & 0.8212 & 0.8633
& 0.9206 & 0.9239 & 0.9427 & 0.9449
& 0.9233 & 0.9178 & 0.8270 & 0.8935 \\

FedRep
& 0.8141 & 0.5503 & 0.8297 & 0.8775
& 0.9179 & 0.8984 & 0.9389 & 0.9423
& 0.9345 & 0.9073 & 0.8118 & 0.9093 \\

FedBN
& 0.8200 & 0.6550 & 0.8287 & 0.8663
& 0.9229 & 0.8973 & 0.9366 & 0.9418
& 0.9341 & 0.9116 & 0.7887 & 0.9032 \\

FedProx
& 0.8248 & 0.6829 & 0.7959 & 0.8173
& 0.8780 & 0.9307 & 0.9334 & 0.9156
& 0.9270 & 0.8955 & 0.8021 & 0.9138 \\

FedAdam
& \textbf{0.8569} & \textbf{0.8332} & \textbf{0.9391} & \textbf{0.8906}
& 0.9287 & 0.9124 & 0.9514 & 0.9511
& 0.9247 & 0.8975 & 0.8124 & 0.9043 \\

SCAFFOLD
& 0.7368 & 0.3799 & 0.7083 & 0.7025
& 0.8715 & 0.8333 & 0.8940 & 0.9055
& 0.7006 & 0.7429 & 0.5701 & 0.5576 \\

pFedMe
& 0.4564 & 0.1201 & 0.2354 & 0.4054
& 0.7233 & 0.8516 & 0.7843 & 0.7732
& 0.4158 & 0.1823 & 0.1654 & 0.1621 \\

Ditto
& \textbf{0.8769} & \textbf{1.0000} & \textbf{1.0000} & \textbf{1.0000}
& 0.9210 & 0.9232 & 0.9343 & 0.9416
& 0.9338 & 0.9026 & 0.8420& 0.9089\\
\bottomrule
\end{tabular}
\label{tab:compound_subtables}
\end{table*}

\begin{table*}[t]
\tiny
\caption{Data harmonization vs. model personalization across classification tasks. ARI = average representative image, SRI = single representative image. Best results are in \textbf{bold}.}
\centering
\setlength{\tabcolsep}{2pt}
\renewcommand{\arraystretch}{1.2}

\begin{tabular}{@{} l | cccc | cccc | cccc @{}}
\toprule
& \multicolumn{4}{c|}{\textbf{Tuberculosis CXR}} 
& \multicolumn{4}{c|}{\textbf{Brain Tumor}} 
& \multicolumn{4}{c}{\textbf{Breast Tumor}} \\

\textbf{Method} 
& C1 & C2 & C3 & C4 
& C1 & C2 & C3 & C4 
& C1 & C2 & C3 & C4 \\
\midrule

\multicolumn{13}{c}{\textit{Baseline methods}} \\ \midrule

Locally centralized
& 0.8462 & 0.6486 & 0.9750 & 0.9883
& 0.9563 & 0.9866 & 0.9690 & 0.9722
& 0.6353 & 0.6770 & 0.7764 & 0.7000 \\

Centralized -- globally tested
& \multicolumn{4}{c}{0.9656}
& \multicolumn{4}{c}{0.9941}
& \multicolumn{4}{c}{0.5872} \\

Centralized -- locally tested
& 0.8768 & 0.6486 & 0.9875 & 0.9881
& 1.0000 & 0.9943 & 0.9911 & 0.9926
& 0.6398 & 0.4413 & 0.6513 & 0.5263 \\

Vanilla FedAvg -- globally tested
& \multicolumn{4}{c}{0.7763}
& \multicolumn{4}{c}{0.9974}
& \multicolumn{4}{c}{0.5466} \\

Vanilla FedAvg -- locally tested
& 0.0000 & 0.0000 & 0.8500 & 0.9883
& 1.0000 & 0.9962 & 0.9956 & 0.9981
& 0.5645 & 0.3976 & 0.6074 & 0.6667 \\

\midrule
\multicolumn{13}{c}{\textit{Harmonization methods}} \\ \midrule

Augmentations
& 0.0901 & 0.0679 & 0.4631 & 0.6385
& 0.9956 & 0.9962 & 0.9956 & 0.9944
& 0.7251 & 0.4158 & 0.5320 & 0.6667 \\

Histogram matching (ARI)
& 0.4937 & 0.4990 & 0.8625 & 1.0000
& 0.9956 & 0.9962 & 0.9823 & 0.9944
& 0.7451 & 0.4483 & 0.5412 & 0.6667 \\

Histogram matching (SRI)
& 0.2536 & 0.3158 & 0.9000 & 1.0000
& 0.9956 & 0.9962 & 0.9603 & 0.9926
& 0.6385 & 0.3750 & 0.5412 & 0.6667 \\

FDA (SRI)
& 0.1317 & 0.0000 & 0.8875 & 0.9542
& 0.9782 & 0.9790 & 0.9294 & 0.9778
& 0.3329 & 0.1832 & 0.3621 & 0.6667 \\

Input level MixStyle
& 0.1898 & 0.0000 & 0.7750 & 1.0000
& 1.0000 & 1.0000 & 0.9734 & 0.9981
& 0.6596 & 0.3750 & 0.5218 & 0.6667 \\

Feature level MixStyle
& 0.0634 & 0.0000 & 0.9250 & 1.0000
& 1.0000 & 1.0000& 0.9823 & 0.9963
& 0.7359 &0.3415 &0.6177 &0.6667 \\

Cycle-GAN
& \textbf{0.5317} & \textbf{0.8158} & \textbf{0.7750} & \textbf{0.9883}
& 0.9825 & 0.9867 & 0.9735 & 0.9648
& 0.6750 & 0.3077 & 0.5053 & 0.1429 \\

CUT
& 0.0000 & 0.0000 & 0.7000 & 0.9883
& 1.0000 & 0.9962 & 0.9647 & 0.9926
& 0.6494 & 0.3333 & 0.2101 & 0.4706 \\

AdaIN
& \textbf{0.5534} & \textbf{0.8434} & \textbf{0.8625} & \textbf{0.8606}
& 0.9956 & 0.9962 & 0.9425 & 0.9889
& 0.4615 & 0.3671 & 0.3081 & 0.4706 \\

ComoGAN
& 0.0000 & 0.0000 & 0.5250 & 1.0000
& 1.0000 & 0.9962 & 0.9734 & 0.9944
& 0.3654 & 0.2751 & 0.3932 & 0.7000 \\

\midrule
\multicolumn{13}{c}{\textit{Personalization methods}} \\ \midrule

Local Finetuning
& 0.0329 & 0.0914 & 0.8678 & 0.8217
& 0.9854 & 0.9760 & 0.9614 & 0.9340
& 0.5874 & 0.4476 & 0.5249 & 0.5682 \\

FedPer
& 0.0634 & 0.0000 & 0.9125 & 0.9883
& 0.9956 & 0.9943 & 0.9911 & 0.9963
& 0.5994 & 0.4551 & 0.6074 & 0.6667 \\

FedRep
& 0.0959 & 0.2041 & 0.9500 & 0.9883
& 0.9956 & 0.9943 & 0.9956 & 0.9926
& 0.6972 & 0.5280 & 0.6275 & 0.4000 \\

FedBN
& 0.0000 & 0.0000 & 0.0000 & 0.0000
& 0.9826 & 0.9522 & 0.9297 & 0.8796
& 0.4639 & 0.5190 & 0.6456 & 0.4545 \\

FedProx
& 0.3223 & 0.0990 & 0.9375 & 0.9883
& 0.9956 & 0.9981 & 0.9735 & 0.9963
& 0.3871 & 0.2298 & 0.4188 & 0.5263 \\

FedAdam
& 0.0000 & 0.0000 & 0.0000 & 0.0000
& 0.0000 & 0.0000 & 0.0000 & 0.0000
& 0.0000 & 0.0000 & 0.0000 & 0.0000 \\

SCAFFOLD
& 0.0000 & 0.0000 & 0.0000 & 0.0000
& 0.9869 & 0.9790 & 0.9691 & 0.9778
& 0.0000 & 0.0000 & 0.0000 & 0.0000 \\

pFedMe
& 0.2547 & 0.1420 & 0.2847 & 0.4875
& 0.0890 & 0.0890 & 0.0305 & 0.1130
& 0.0000 & 0.0000 & 0.0000 & 0.0000 \\

Ditto
& 0.7841 & 0.6486 & 1.0000 & 0.9268
& 0.9956 & 0.9943 & 0.9956 & 0.9926
& 0.6446 & 0.3191 & 0.5849 & 0.6667 \\
\bottomrule
\end{tabular}
\label{tab:compound_subtables_classification}
\end{table*}

For colon polyp segmentation, where inter-client variation is large and primarily structural (e.g., shape and geometry), personalization methods consistently outperform harmonization, highlighting the need to adapt models to client-specific structures. In contrast, for skin lesion segmentation, where variation is more limited and largely appearance-driven, harmonization methods (e.g., MixStyle, SRI, ComoGAN) perform comparably to personalization. Breast tumor segmentation follows a similar pattern, with both strategies achieving competitive performance, suggesting that when structural differences are moderate, neither approach has a clear advantage.

A similar trend is observed in classification tasks. For brain tumor classification, where inter-site variation is minimal, all methods reach near-ceiling performance, and neither harmonization nor personalization provides a clear benefit. Breast tumor classification shows comparable behavior, with both approaches yielding similar results across clients. In contrast, for tuberculosis CXR classification, where variation is dominated by appearance differences, harmonization consistently outperforms personalization by reducing input variability.

Overall, when inter-client differences are small or moderate, both strategies perform similarly regardless of task. When differences are large, the dominant type of variation determines the more effective approach: personalization is better suited for structurally heterogeneous segmentation tasks, whereas harmonization is more effective for appearance-driven classification tasks.

\section{Conclusions \& Future works}
\label{sec:conclusions}
The results reveal a conditional trade-off rather than a strictly task-dependent pattern. When inter-client differences are large and primarily structural, as in colon polyp segmentation, personalization outperforms harmonization by adapting to client-specific geometry and context. In contrast, when variation is dominated by appearance and intensity differences, as in tuberculosis CXR classification, harmonization is more effective by reducing input variability and supporting a shared global model. However, this trend does not hold universally. When inter-client differences are small or moderate, both strategies perform similarly, as observed in skin lesion and breast tumor segmentation, as well as brain and breast tumor classification, where neither approach offers a clear advantage. Overall, the effectiveness of harmonization versus personalization is governed by the type and magnitude of dataset heterogeneity rather than the task alone. In practice, federated medical imaging systems should assess cross-site variation before selecting a strategy, rather than assuming a fixed preference. A key limitation is the assumption that inter-client heterogeneity is known a priori, whereas in practice it is unknown and dynamic. Future work should develop metrics to characterize heterogeneity directly from data and enable adaptive frameworks that combine harmonization and personalization at multiple levels. Extending this analysis to non-stationary settings and incorporating causal modeling are important directions. Standardized benchmarks and evaluation of performance–privacy–communication trade-offs are also needed.

\begin{credits}


\subsubsection{\discintname}
The authors have no competing interests to declare that are
relevant to the content of this article.

\mycomment{
It is now necessary to declare any competing interests or to specifically
state that the authors have no competing interests. Please place the
statement with a bold run-in heading in small font size beneath the
(optional) acknowledgments\footnote{If EquinOCS, our proceedings submission
system, is used, then the disclaimer can be provided directly in the system.},
for example: The authors have no competing interests to declare that are
relevant to the content of this article. Or: Author A has received research
grants from Company W. Author B has received a speaker honorarium from
Company X and owns stock in Company Y. Author C is a member of committee Z.}
\end{credits}
%
%
%
%

\end{document}